# Assessment of Using Synthetic Data in Brain Tumor Segmentation

Aditi Jahagirdar
*Department of Information & Technology*
*IU International University of Applied Sciences*
Bangalore, India
aditi.jahagirdar@iu-study.org

Prof. Dr. Sameer Joshi
*Department of Management & Business Studies*
*IU International University of Applied Sciences*
Erlangen, Germany
sameer.joshi@iu.org

*Abstract*—Manual brain tumor segmentation from MRI scans is challenging due to tumor heterogeneity, scarcity of annotated data, and class imbalance in medical imaging datasets. Synthetic data generated by generative models has the potential to mitigate these issues by improving dataset diversity. This study investigates, as a proof of concept, the impact of incorporating synthetic MRI data, generated using a pre-trained GAN model, into training a U-Net segmentation network. Experiments were conducted using real data from the BraTS 2020 dataset, synthetic data generated with the medigan library, and hybrid datasets combining real and synthetic samples in varying proportions. While overall quantitative performance (Dice coefficient, IoU, precision, recall, accuracy) was comparable between real-only and hybrid-trained models, qualitative inspection suggested that hybrid datasets, particularly with 40% real and 60% synthetic data, improved whole tumor boundary delineation. However, region-wise accuracy for the tumor core and the enhancing tumor remained lower, indicating a persistent class imbalance. The findings support the feasibility of synthetic data as an augmentation strategy for brain tumor segmentation, while highlighting the need for larger-scale experiments, volumetric data consistency, and mitigating class imbalance in future work.

## I. Introduction

A brain tumor is an abnormal growth of cells in the brain. Non-cancerous tumors such as glioma are also called low-grade tumors, which are structured and grow slowly. Cancerous tumors such as glioblastoma are structurally non-uniform, grow rapidly and are called high-grade tumors. As it consists of active cancer cells, it is classified as a malignant tumor [1]. Patients with low-grade glioma who can survive for 10 years is 57%. Only 8% of chronic cases of malignant tumors, such as high-grade glioblastoma and metastasis, survive for 2.5 years, and the survival rate for 10 years is even lower at just 2% [2]. Incidence rates in the year 2022 [3] constitute 322,000 new cases globally of brain and central nervous system tumors. It is projected that the cases will increase by 47% (accounting for 474,000 new cases per year) by the year 2045, when compared to 2022 [3]. These anticipated rates emphasise the need for early and accurate diagnosis, which is necessary to devise and start the appropriate course of treatment with the least delay, so that the lives of patients can be saved.

In this study, MRI scans are utilised for training the model to identify granular details of the tumor. Brain MRI scans are divided into 4 key regions of tumor components which can be found in available datasets [4] as Whole Tumor (WT) which includes all abnormal cells to estimate the overall size of the tumor, Enhancing Tumor (ET) represents which section of the tumor is growing rapidly compared to surrounding tissues, Tumor Core (TC) consists of ET and dead cells and Background (BG) are the healthy brain tissues that are not part of the tumor. In multimodal MRI sequences, each of the modalities captures different regions of the brain by using contrast mechanisms. Manually diagnosing a brain tumor is a labour-intensive process where the reliability of the diagnosis differs based on the observer's skills. It is difficult to identify smaller lesions in the context of brain anatomy and identify boundaries when there are similar tissue structures. Hence, automation of medical imaging analysis can streamline the process to improve the consistency and accuracy of the results.

The accuracy of brain tumor segmentation is highly impacted due to dataset issues like scarcity of data, poor quality images, inconsistent annotations and class imbalance. Another prominent challenge is the unavailability of diverse data, which leads to overfitting of the model and impacts its generalization ability. The predictions are biased due to class imbalance, which can lead to inaccurate diagnoses. To summarize, the dataset issues adversely impact the robustness of the model and produce unreliable results and hence require prompt attention.

The objective of this study is to assess the usage of synthetic data as an augmentation technique to address the challenges in brain tumor segmentation arising from dataset issues. It should be noted that in the context of this study, segmentation of tumor refers to glioma, which is a cancerous brain tumor originating in the brain or spinal cord. Glioma accounts for 30% of all brain tumors [5]. This research will answer how brain tumor segmentation performance is impacted if synthetic data is used in the model training. Is the segmentation accuracy affected adversely, improved, or remains the same? A novel pipeline is designed to evaluate the model's performance on different combinations of hybrid datasets. It should be noted that the study focuses on addressing the dataset issues only, and any other limitations of segmentation modelling, such as computational efficiency and non-standardized imaging protocols, are out of scope for this research.

## II. Literature Review

A theoretical and evidence-based literature review was conducted using meta-analysis and survey studies. To keep the focus on contributing to the advancement in the topic of study, the last 5 years of the latest and related work were referred from peer-reviewed journals, conference papers, government-funded or university-supported published work and some of the early-stage papers. The major databases that were queried for the literature review were ScienceDirect, DOI, arXiv, Springer and IEEE Xplore.

The focus of this study is on image segmentation, which is a technique that divides an image into segments based on intensity, colour, texture, and other factors, depending on its use case and the characteristics of the data. The segmentation model is based on a U-Net architecture, which is a Convolutional Neural Network (CNN) and an excellent example where features are extracted in the contraction path and a segmented image is reconstructed in the expansion path. U-Net can achieve high performance with limited datasets, which is why it is popularly used in medical image segmentation. Despite being an efficient architecture, the data-related factors discussed in the introduction affect the performance of the U-Net model and the accuracy of predictions. To address these challenges, data augmentation techniques are employed, which are described in the figure below. To explore generative modeling as a data augmentation technique, Generative Adversarial Networks (GAN) architecture is used to generate synthetic data in this research.

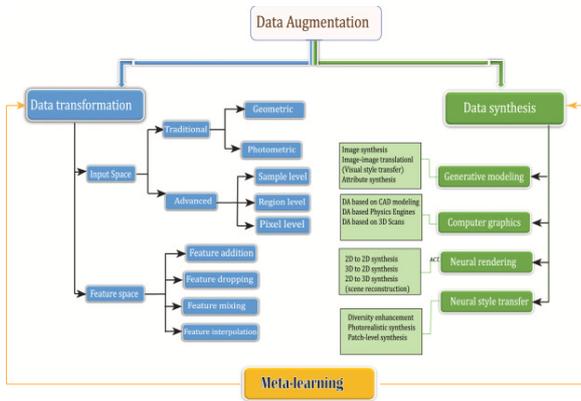

Fig. 1. Source: [6] A comprehensive overview of data augmentation techniques.

A few of the latest work in brain tumor segmentation includes CFNet (coarse-to-fine feature fusion network), which is designed to segment smaller lesion regions accurately [7]. The modified U-Net with attention gate has integrated an attention gate mechanism with the U-Net architecture [8]. MSDMAT-BTS (Multi-scale diffusion model and attention mechanism) addresses the challenges of capturing both local and global characteristics by current diffusion models using a multi-scale diffusion model with an attention mechanism [9]. Segment Anything Model (SAM) uses fine-tuning strategies like parameter-efficient learning to improve the model's performance [10]. Lightweight architectures such as T-Net design [11] and U-Lite [12] are useful to overcome the constraints in computational resources. Overall, the theme interpreted from recent research on brain tumor segmentation indicates that the creation of novel networks using attention mechanisms in segmentation architectures such as diffusion models and U-Net models has improved the performance and segmentation accuracy. The issues with data diversity and computational complexity are still prevalent and need to be addressed. The survey [13] discusses the problem of data scarcity and how synthetic images address it. GANs are the most widely used architecture for generating synthetic data, while other techniques, such as diffusion models, transformers and RNNs are also advancing. Synthetic data generated using the Enhancement GAN (EnhGAN) framework [14] demonstrated improved accuracy of region-wise segmentation by minimising the overlapping distributions of different brain tissue types, which confuse the model.

Evaluation of the GAN and diffusion model for generating synthetic images and then using it in brain tumor segmentation networks, U-Net, and Swin transformer observed that the segmentation network slightly performed better on real datasets compared to synthetically generated data [15]. VarVit-GAN is used for generating T1 contrast-enhanced images to address data scarcity and missing data issues [16]. BrainGAN framework [17] used images generated by Vanilla GAN and DCGAN to train the tumor classification model using ResNet152V2, MobileNetV2 and CNN architectures. ResNet152V2 achieved comparatively higher performance. Several comparative studies have analysed and refined the evaluation metrics for synthetic data, which further helps to improve the quality of data generated.

GAN is noted as the most popular architecture used for generating high-quality, realistic images. Based on the conducted literature review, further scope for improvement is identified in improving the generalization ability of segmentation models by using synthetic data, which addresses the dataset issues. Both segmentation models and GAN variants shall focus on modelling the architecture such that it is computationally efficient and trains faster. Class imbalance issues should be considered to enhance the accuracy for better boundary detection and fine-grained feature segmentation. Especially in the case of medical image analysis, careful attention is required to maintain ethical and privacy protocols.

Over the past decade, significant advancements have been observed in medical imaging segmentation and generative modelling. Separate studies are available on using GAN for synthetic image generation and brain tumor segmentation. However, there is limited research available on the usage of GAN-generated data to create hybrid datasets, which are a combination of real and synthetic data and use them for training brain tumor segmentation models, which forms the scope of this research.

III. METHODOLOGY

This work is based on the hypothesis that the diagnostic accuracy of the brain tumor segmentation model is not impacted adversely and may improve as well if synthetic data is used for data augmentation in the training process. This hypothesis is evaluated through an experimental study using varied datasets, which is divided into two key parts: the first one being the generation of synthetic data, and the second part being the segmentation of a brain tumor from an MRI image. The real dataset used for training the U-Net model is the publicly available BraTS2020 dataset [18].

Datasets used were categorised as real-only data, hybrid data (consisting of different combinations of synthetic data) and synthetic-only data. The Generative Adversarial Network (GAN) [19] is used for artificially generating data, and the U-Net architecture [20] is used for the segmentation task. Experimental setup, computing environment, tools and frameworks and dataset preparation are presented below.

*A. Experimental setup*

To conduct the experiments discussed above, a pipeline shown in the figure below is designed that explains the process of training a brain tumor segmentation model using different types of datasets. Firstly, the synthetic dataset is generated using a Generative Adversarial Network (GAN), which is then processed so that it is compatible with the real dataset. This compatibility is required for the ease of creating a hybrid dataset, which is a combination of real and synthetic data in different proportions (combination 1: 60% synthetic data, combination 2: 80% synthetic data and combination 3: 90% synthetic data).

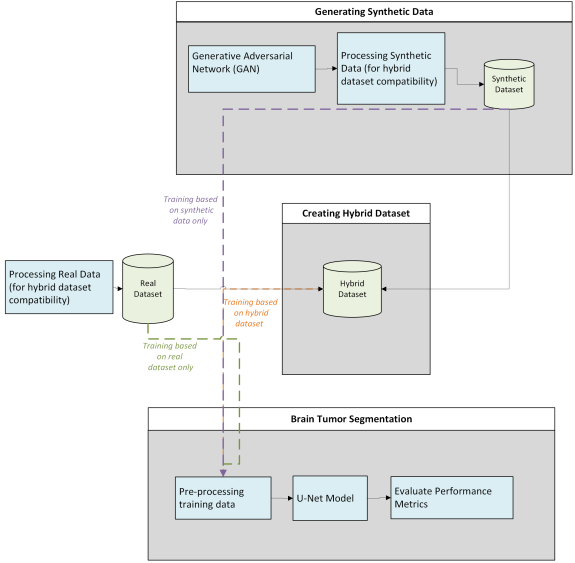

Fig. 2. Pipeline for training a brain tumor segmentation model.

The U-Net architecture is used for brain tumor segmentation, and it is trained on these different datasets. The performance metrics are then evaluated to assess the impact of using synthetic data on the model's performance for the segmentation task.

*B. Computing environment*

Python (version 3.7.10) was used as a programming language throughout the pipeline. Medigan [21] library was used for the generation and processing of synthetic images. Hybrid data set combinations were created manually by visual inspection of the data. For the brain tumor segmentation model, PyTorch (1.9.1+cpu) [22] and Torchvision (0.10.1+cpu) [23] frameworks were used.

The '00007_INPAINT_BRAIN_MRI' model of the medigan library is used in the experiments for generating brain tumor images with masks [21]. This tumor inpainting model is trained on the BraTS 2018 dataset [24] and generates 4 modalities (t1, t1c, flair and t2) of brain MRI images along with their corresponding binary and tumor-grade masks. The binary mask only identifies if the tumor is present or not. Whereas tumor grade masks assign different labels to different subregions of tumor, such as 'Label 0' for background (BG), 'Label 1' for Necrotic and non-enhancing tumor core (NCR/NET), 'Label 2' for Peritumoral edema (ED) and 'Label 4' for GD-enhancing tumor (ET). A total of one hundred samples were generated with one inpaint per sample. The remaining parameters used in the generator were as provided in the example usage of this model. The generated output images are 2-dimensional grayscale images of size 256x256. These images in PNG (Portable Network Graphic) format were stored on a local machine, constituting 10.3 MB of disk space.

*C. Dataset preparation*

The characteristics of 100 sample-sized real and synthetic training data can be referred to the table below:

TABLE 1. CHARACTERISTICS OF REAL AND SYNTHETIC DATA

| Characteristics | Synthetic data (generated by medigan) | Real data (BraTS 2020) |
|---|---|---|
| Image dimensions | 2-D | 3-D |
| Image size | 256x256 | 240x240x155 |
| Number of images per patient | 6 (4 multimodal MRI images, binary segmentation mask, tumor grade segmentation mask) | 5 (4 multimodal MRI images, tumor grade segmentation mask) |
| Number of channels | 3 (RGB) | 1 (Grayscale) |
| File format | PNG (.png) | NIFTI (.niii.gz) |

Synthetic images generated by medigan library were grouped patient-wise, excluding the binary mask and the data format was processed to match with real dataset. Images from both the real dataset and the synthetic dataset were converted to JPG format, and finally, 100 samples were selected for each dataset by visually inspecting the quality of the images.

To segment the tumor, a 2D U-Net model is used where the image is gray-scaled, resized to 240x240, multimodal images were stacked, transposed and normalised to ensure consistent size, preserve spatial alignment, learn comprehensive feature representation and stabilise training. The hyperparameters which were tested and tuned during training of the model include a batch size of 4, accumulation steps of 4, trained for 50 epochs with a learning rate of $5e^{-4}$ using the Adam optimizer and a custom loss function, which was a combination of Dice loss and Binary Cross Entropy (BCE) loss. If the loss on the validation dataset did not improve for 5 epochs, then early stopping [25] was triggered. The performance metrics used were precision, recall, accuracy, dice coefficient and Jaccard index (IoU – Intersection over Union).

IV. RESULTS AND DISCUSSION

The model trained on a real dataset is the baseline model, consisting of high-resolution images, but some MRI scans have exhibited artifacts like blurring and distortions. Also, no significant impact of processing performed on a dataset from 3D to 2D was observed on the metrics, apart from the segmentation quality. This operation was required for compatibility in the creation of a hybrid dataset.

Training of the hybrid dataset model, as observed in the figure below, showed a similar trend to that of a model trained on real data in terms of stabilizing around the same number of epochs with comparable loss and metrics.

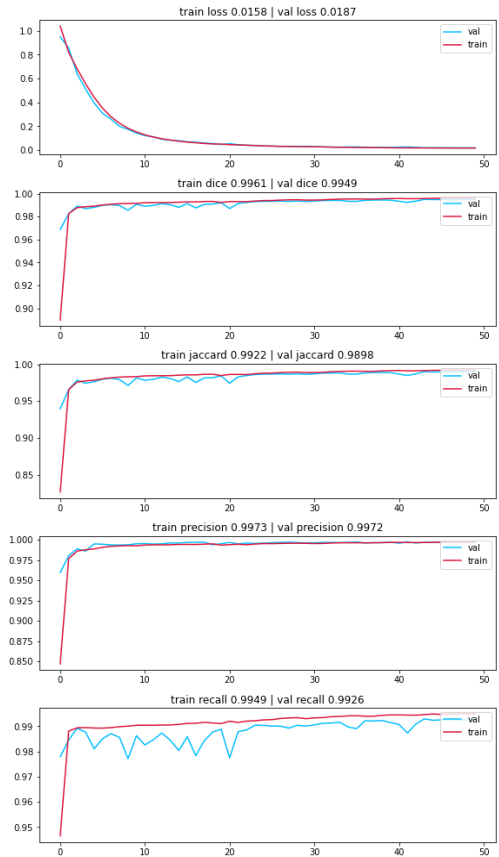

Fig. 3. Training history (metrics versus epochs) of the segmentation model on a hybrid dataset with 60% synthetic data.

While the metrics for the real-only dataset demonstrated high accuracy for each of the regions, the predicted masks from the validation output are unable to capture the features as visually observed in the figure below.

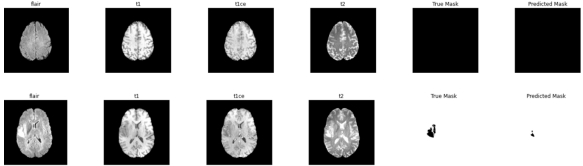

Fig. 4. Validation output of the real dataset. From left to right: FLAIR, T1, T1CE, T2, ground truth mask and predicted mask.

The model trained on a combination of 40% real dataset and 60% synthetic dataset struggled to segment precise regions of tumor, but whole tumor segmentation was visually better than the model trained on the real-only dataset, as observed in the figure below. For the hybrid dataset combination of 20% real dataset and 80% synthetic dataset, the class imbalance can be observed as becoming more prominent with an increasing proportion of synthetic data. For a hybrid dataset consisting of 10% real data and 90% synthetic data, region-wise segmentation accuracy was improved compared to other hybrid dataset combinations. The accuracy of validation data from the synthetic-only dataset was 93.1% for the whole tumor (WT), 54.1% for the tumor core (TC), 65.8% for the enhanced tumor (ET) and 99.9% for background (BG). The segmentation quality was visibly improved for the synthetic-only dataset compared to the real-only dataset. This implies the need for additional pre-processing to refine the segmentation masks from the real dataset.

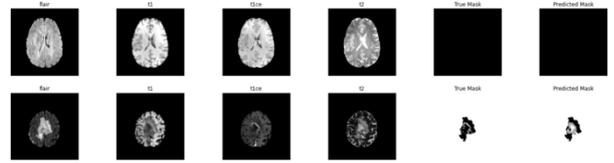

Fig. 5. Validation output of the hybrid dataset with 60% synthetic data. From left to right: FLAIR, T1, T1CE, T2, ground truth mask and predicted mask.

The accuracy of hybrid dataset variants is tabulated below where WT represents the Whole Tumor region, TC is the Tumor core, ET is the Enhanced Tumor and BG is the background region.

TABLE 2. ACCURACY OF HYBRID DATASETS

| % of Synthetic data | WT | TC | ET | BG |
| --- | --- | --- | --- | --- |
| 60% | 93.2% | 46.7% | 69.7% | 99.9% |
| 80% | 91.3%, | 37% | 69.6% | 99.9% |
| 90% | 92.7% | 46.8% | 67.9% | 99.9% |

The values in all test cases implied good-quality segmentation performance, excellent overlap between predicted and ground truth masks and the model's proficiency in detecting true positives and rarely predicting any false positives. The accuracy for models trained using synthetic data (in different proportions) reflected class imbalance, with lower accuracy for tumor core (TC) and enhanced tumor (ET) regions. However, the accuracy of whole tumor segmentation was comparatively better with an average value of 92.6% for the test cases using synthetic data.

The hybrid dataset combination which performed well compared to other combinations was the one with 40% real data and 60% synthetic data. As the training data was getting skewed with one source of the dataset, the region-wise segmentation accuracy showed a slight decline. Hence, a careful balance shall be identified between real and synthetic data while designing such combinations, and it would also depend on the characteristics of the datasets used in the process. Segmentation quality on validation data was better in the models trained using synthetic data compared to the validation output images of the model trained on real-only datasets, which indicated the requirement for better pre-processing of segmentation masks from real data.

Overall, the experimental results were conclusive that high-quality tumor segmentation can be achieved by using synthetic data as a data augmentation technique. If the class imbalance problem can be addressed in the generated synthetic dataset, then the performance of the hybrid dataset is expected to improve further. Hence, the hypothesis stating that the diagnostic accuracy is not impacted adversely and may improve with the usage of synthetic data as an augmentation technique in tumor segmentation is accepted after a careful assessment of the experimental findings.

## V. CONCLUSION

This study evaluated the feasibility of using GAN-generated synthetic MRI data to augment real datasets for brain tumor segmentation with a U-Net architecture. Across real-only, synthetic-only, and hybrid datasets, segmentation performance in whole tumor regions was consistently high, with hybrid datasets, particularly with 40% real and 60% synthetic data, achieving comparable quantitative metrics to real-only training and, in some cases, improved qualitative boundary delineation. Nevertheless, lower accuracy in the tumor core and enhanced tumor regions reflected persistent class imbalance, and the conversion of real 3D volumes to 2D slices may have reduced segmentation mask quality by limiting spatial context.

The study was intentionally small-scale, using 100 samples per dataset variant, which is sufficient for proof-of-concept but not for definitive benchmarking. Another limitation was the reliance on a pre-trained GAN model from the medigan library, which may introduce domain overlap with the segmentation dataset. Future research should explore self-designed 3D generative models to minimise preprocessing, handle class imbalance during training, and reduce dependency on pre-trained models trained on similar datasets. Expanding datasets with diverse sources such as the Cancer Imaging Archive, Kaggle, and clinical repositories will improve generalization and reduce bias.

By addressing these factors and considering newer generative approaches such as diffusion models, attention mechanisms and enhanced CNN architecture, synthetic data augmentation has the potential to become a robust and scalable strategy for improving brain tumor segmentation performance in clinical and research applications.